\documentclass[10pt, journal, compsoc]{IEEEtran}

\usepackage{times}  
\usepackage{helvet} 
\usepackage{courier}  
\usepackage[hyphens]{url}  
\usepackage{graphicx} 
\usepackage{amsmath}
\usepackage{nicefrac}
\usepackage[super]{nth}
\usepackage{bm}
\usepackage{upgreek}
\usepackage{caption}
\usepackage{algpseudocode}
\usepackage{amssymb}
\usepackage{xcolor}
\usepackage{enumitem}
\usepackage{amsthm}
\usepackage{multirow}
\usepackage{setspace}
\usepackage[ruled,linesnumbered]{algorithm2e}
\usepackage{mathtools}
\usepackage{bm}
\usepackage{nicefrac}
\usepackage{lipsum}

\algnewcommand\algorithmicforeach{\textbf{for each}}
\algdef{S}[FOR]{ForEach}[1]{\algorithmicforeach\ #1\ \algorithmicdo}

\newcommand{\E}{\mathbb{E}}

\newcommand{\bbmxt}{\bar{\bm{x}}_t}

\title{Faster Federated Learning with Decaying \\ Number of Local SGD Steps}

\begin{document}

\author{Jed Mills, Jia Hu, Geyong Min
	\IEEEcompsocitemizethanks{
		\IEEEcompsocthanksitem{ J. Mills, J. Hu and G. Min are with the 
		Department of Computer Science, University of Exeter, EX4 4QF, United Kingdom. 
		E-mail: \{jm729, j.hu, g.min\}@exeter.ac.uk. Corresponding authors: Jia Hu, Geyong Min.} 
}}

\IEEEtitleabstractindextext{%
\begin{abstract}
In Federated Learning (FL) client devices connected over the internet collaboratively train a machine learning  model without sharing their private data with a central server or with other clients. The seminal Federated Averaging (FedAvg) algorithm trains a single global model by performing rounds of local training on clients followed by model averaging. FedAvg can improve the communication-efficiency of training by performing more steps of Stochastic Gradient Descent (SGD) on clients in each round. However, client data in real-world FL is highly heterogeneous, which has been extensively shown to slow model convergence and harm final performance when $K > 1$ steps of SGD are performed on clients per round. In this work we propose decaying $K$ as training progresses, which can jointly improve the final performance of the FL model whilst reducing the wall-clock time and the total computational cost of training compared to using a fixed $K$. We analyse the convergence of FedAvg with decaying $K$ for strongly-convex objectives, providing novel insights into the convergence properties, and derive three theoretically-motivated decay schedules for $K$. We then perform thorough experiments on four benchmark FL datasets (FEMNIST, CIFAR100, Sentiment140, Shakespeare) to show the real-world benefit of our approaches in terms of real-world convergence time, computational cost, and generalisation performance.
\end{abstract}

\begin{IEEEkeywords}
Federated Learning, Deep Learning, Edge Computing, Computational Efficiency.
\end{IEEEkeywords}}

\maketitle

\section{Introduction}
\IEEEPARstart{F}{ederated} Learning (FL) is a recent distributed Machine Learning (ML) paradigm that aims to collaboratively train an ML model using data owned by clients, without those clients sharing their training data with a central server or other participating clients. Practical applications of FL range from `cross-device' scenarios, with a huge number of unreliable clients each possessing a small number of samples, to `cross-silo' scenarios with fewer, more reliable clients possessing more data \cite{AdvancesinFL}. FL has huge economic potential, with cross-device tasks including mobile-keyboard next-word prediction \cite{FLMobileKeyboard}, voice detection \cite{FLKeywordSpotting}, and even as proof-of-work for blockchain systems \cite{ProofOfFL}. Cross-silo tasks include hospitals jointly training healthcare models \cite{FLInMedicine} and financial institutions creating fraud detectors \cite{FLFraud}. FL has been of particular interest for training large Deep Neural Networks (DNNs) due to their state-of-the-art performance across a wide range of tasks.

Despite FL's great potential for privacy-preserving ML, there exist significant challenges to address before FL can be more widely adopted at the network edge. These include:

\begin{itemize}[leftmargin=*]
\item{\textit{Heterogeneous client data:} each client device generates its own data, and cannot share it with any other device. The data between clients is therefore highly heterogeneous, which has been shown theoretically and empirically to harm the convergence and final performance of the FL model.}

\item{\textit{High communication costs:} many FL algorithms operate in rounds that involve sending the FL model parameters between the clients and the coordinating server thousands of times. Considering the bandwidth constraints of wireless edge clients, communication represents a major hindrance to training. }

\item{\textit{High computation costs:} training ML models has a high computational cost (especially for modern DNNs with a huge number of parameters). FL clients are typically low-powered (often powered by battery), so computing the updates to the FL model is a substantial bottleneck. }

\item{\textit{Wireless edge constraints:} clients are connected to the network edge and can range from modern smartphones to Internet-of-Things (IoT) devices. They are highly unreliable and can leave and join the training process at any time. }
\end{itemize}

\noindent To address some of the above challenges, McMahan \textit{et al.} proposed the Federated Averaging (FedAvg) algorithm \cite{FedAvgPaper}. FedAvg is an iterative algorithm that works in communication rounds, where in each round clients download a copy of the `global model' to be trained, perform $K$ steps of Stochastic Gradient Descent (SGD) on their local data, then upload their models to the coordinating server, which averages them to produce the next round's global model. Therefore, FedAvg works similarly to distributed-SGD (dSGD) as used in the datacentre, but more than one gradient is calculated by clients per communication round. Using $K > 1$ local steps improves the per-round convergence rate compared to dSGD (hence saving on communication), and FedAvg only requires a fraction of all clients to participate in each round, mitigating the impact of unreliable clients and stragglers.

Increasing $K$ improves the convergence rate (in terms of communication rounds) of FedAvg, however it has been demonstrated that it comes at the cost of harming the minimum training error and maximum validation accuracy than can be achieved, especially when client data is heterogeneous \cite{AdvancesinFL}, and large values of $K$ show diminishing returns for convergence speed. Therefore the total amount of computation performed to reach a given model error can be significantly greater compared to datacentre training, leading to concerns over the energy cost of FL  \cite{FLSavePlanet}. Furthermore, the computation time on low-powered FL clients is not negligible, so improving communication-efficiency by using larger $K$ can lead to a long training procedure \cite{TowardsEffSched,CommEffFLIoT}.

The primary reason behind the performance degradation with increasing $K$ is `client-drift' \cite{SCAFFOLD}: as the data between clients is non-Independent and Identically Distributed (non-IID), the minimum point(s) of each client's objective will be different. During local training, client models diverge (drift) towards their disparate minimisers, and the average of these disparate models may not have good performance. The extent of client-drift has been shown theoretically to be proportional to the level of heterogeneity between client data, the client learning rate ($\eta$), and $K$ \cite{SCAFFOLD,ConvergenceFedAvgNonIID}. 

One theoretically-justified method of addressing the problem of client-drift is to reduce $\eta$ during training. Intuitively, if the learning rate is smaller then client models can move less far apart during the local update. Previous works have shown that decaying $\eta$ is required for the error of the global model to converge to $0$. In this paper, we propose instead decaying $K$ to achieve a similar goal. Decreasing $K$ addresses client-drift whilst reducing the real-time and computational cost of each FedAvg round. We show in experiments using benchmark FL datasets that decaying $K$ can match or outperform decaying $\eta$ in terms of time to converge to a given error, total computational cost, and maximum validation accuracy achieved by the model. The main contributions of this paper are as follows:

\begin{itemize}[leftmargin=*]
\item{We analyse the convergence of FedAvg when using a decreasing value of $K$ for strongly-convex objectives, which provides novel insight into the constraints on $K$ and $\eta$, and intuitively demonstrates why and demonstrates the impact of $K > 1$ on convergence.}

\item{We derive the optimal value of $K$ for any point during the training runtime, and use this optimal value to propose two theoretically-motivated approaches for decaying $K$ based either on the communication round or the relative FL model error. We also use the analysis to derive the optimal value of $\eta$ for later comparison.}

\item{We perform extensive experiments using four benchmark FL datasets (FEMNIST, CIFAR100, Sentiment140, Shakespeare) to show that the proposed decaying-$K$ scheme can reduce the amount of real-time taken to achieve a given model error, as well as improving final model validation performance.}

\item{We present a further practical heuristic for decaying $K$ based on training error which also shows excellent performance in terms of improving the validation performance of the FL model on the four benchmark datasets.}

\end{itemize}

\noindent The rest of this paper is organised as follows: in Section 2 we cover related works that analyse the convergence properties of FedAvg, algorithms designed to address client-drift, and relevant developments in datacentre-based training; in Section 3 we formalise the FL training objective, analyse the convergence of FedAvg using our proposed decaying-$K$ schedule, derive the optimal value of $K$ during training, and use this to motivate three $K$-decay schemes; in Section 4 we present an experimental evaluation of the proposed schemes; and in Section 5 we conclude the paper.

\section{Related Work}
In this section, we cover works that study the theoretical convergence properties of FedAvg (and related algorithms) and client-drift, algorithms that improve the convergence of FL, and works that study related problems in the datacentre setting.

\subsection{Analysis of FedAvg}
There has been significant research efforts in theoretically analysing the convergence of FedAvg. Li \textit{et al.} \cite{ConvergenceFedAvgNonIID} proved a convergence rate of $\mathcal{O}(1/T)$ (where $T$ is equal to the number of total iterations, rather than communications rounds) on strongly-convex objectives. Their analysis suggests that an optimal number of local steps ($K$) exists to minimise the number of communication rounds to reach $\epsilon$-precision, and the authors highlighted the need to decay the learning rate ($\eta$) during training. Non-dominant convergence in terms of total iterations remains an open problem within FL.  Karimireddy \textit{et al.} \cite{SCAFFOLD} added a server learning rate to FedAvg to prove its convergence for nonconvex objectives. Charles and Konecn\'{y} \cite{ConvAccTrades} analysed the convergence of Local-SGD methods (including FedAvg) for quadratic objectives to gain insights into the trade-off between convergence rate and final model accuracy. Malinovsky \textit{et al.} \cite{SGDToFixedPoint} generalised Local-SGD methods to generic fixed-point functions to analyse the effect of $K$ on the $\epsilon$-accuracy. Yang \textit{et al.} \cite{AchievingLinearSpeedup} were the first to achieve linear speedup in terms of number of participating workers for FedAvg on nonconvex objectives.  However, when considering partial worker participation (which is a key element of the FL scenario), their analysis does not show speedup with respect to $K$. Previous works have also analysed the convergence of FedAvg from perspectives such as minimising the total energy cost and optimal resource allocation \cite{FLOverWirelessNetworks}.

While the above works analyse the convergence of FedAvg in terms of total iterations and/or communication rounds, the runtime of FedAvg is affected by multiple factors: model convergence rate, total number of local SGD steps, communication bandwidth, model size, and the compute power of client devices. These factors must all be considered if the objective is to improve the runtime of FedAvg, as we do in this work.

\subsection{Novel FL Algorithms}
Due to FL's long training times and the challenging distributed edge environment, a large number of novel algorithms have been designed to improve the convergence rate of FedAvg. Li \textit{et al.} \cite{FedProx} proposed FedProx, which adds a proximal term to client objectives penalising the distance to the current global model. Karimireddy \textit{et al.} \cite{SCAFFOLD} added Stochastic Variance-Reduced Gradients (SVRG) to FedAvg in SCAFFOLD, demonstrating significant speedup on popular FL benchmarks. Empirical convergence rates have also been improved by adding adaptive optimisation to FedAvg both locally \cite{CommEffFLIoT} and globally \cite{Mime}. Adaptive optimisation has also been implemented during the server-update of FedAvg \cite{AdaptiveFedOpt}, which can accelerate convergence without increasing the per-round communication or computation costs for clients. A recent survey covering developments in FL algorithms and their relation to the communications properties of FL is given in \cite{ClientSideOptSurvey}.

The above algorithms can be considered variants of FedAvg in that they perform rounds of local training and model averaging. Our proposed method of decaying the number of local steps during training could in principle be used with any FedAvg variant, which is a potential avenue for future research.

\subsection{Datacentre Training}
Distributed training in the datacentre shares similarities with the FL scenario, and there exists a substantial body of work studying datacentre training. The classic datacentre-based algorithm is distributed-SGD (dSGD), where nodes each compute a single (often very large) minibatch gradient and send it to the parameter server for aggregation.  Woodworth \textit{et al.} \cite{IsLocalSGDBetter} proved for quadratic objectives that Local-SGD methods (which perform multiple steps of SGD between aggregations) converge at least as fast as dSGD (in terms of total iterations), but that Local-SGD does not dominate for more general convex problems. Similarly, Wang \textit{et al.} \cite{UnifiedFrameworkCommEff} unified the analysis of various algorithms related to Local-SGD, covering different communication topologies and non-IID clients, achieving state-of-the-art rates for some settings. Lin \textit{et al.} \cite{DontUseLargeMinibatches} presented a thorough empirical study showing that local-SGD methods generalise better than large-batch dSGD, motivating their approach of switching from dSGD to local-SGD during the later stages of training. Another approach to improve the generalisation performance of large-batch dSGD are `extra-gradient' methods that compute gradient updates after a step of SGD before applying them to the global model \cite{ExtrapLargeBatchTraining}.

While these works present methods that variously improve runtime or generalisation performance, their findings cannot be directly applied to FL. From a theoretical perspective, the primary differences are FL's highly non-IID clients and very low per-round participation rates (which can be as low as $0.1\%$ \cite{FLAtScale}). FL client also have much lower communication bandwidth and computational power compared to datacentre compute nodes.

\section{FedAvg with Decaying Local Steps}
We now formally describe the FL optimisation problem, theoretically analyse the convergence of FedAvg with a decaying number of local SGD steps, and present theoretically-motivated schedules based upon the analysis.

\subsection{Problem Setup}
In FL there are a large number of clients that each possess a small number of local samples. The objective is to train model $\bm{x}$ to minimise the expected loss over all samples and over all clients, namely:
\begin{equation}
F(\bm{x}) = \sum_{c=1}^{C} p_c f_c(\bm{x}) = \sum_{c=1}^{C} p_c \left[ \sum_{n=1}^{n_c} f(\bm{x}, \xi_{c,n}) \right],
\end{equation}
where $C$ is the total number of FL clients, $p_c$ is the fraction of all samples owned by client $c$ (such that $\sum_{c=1}^{C} p_c = 1$), $f$ is the loss function used on clients, and $\{ \xi_{c,1}, \cdots, \xi_{c,n_c} \}$ represent the training samples owned by client $c$. 

To minimise $F(\bm{x})$ in a communication-efficient manner, FedAvg (presented in Algorithm 1) performs multiple steps of SGD on each client between model averaging. FedAvg operates in communication rounds, where in each round $r$ a subset of clients $\mathcal{C}_r$ download the current global model $\bm{x}_r$ (line 5), perform $K_r$ steps of SGD on their local dataset (lines 6-8), and then upload their new models to the coordinating server (line 9). The server averages the received client models to produce the next round's global model (line 11).

FedAvg is typically described and analysed as selecting a subset of clients uniformly at random to participate in each communication round (line 3). However in real-world FL deployments, clients generally do not participate uniformly at random due to their behaviour, communication and compute capabilities. The non-uniform participation of FL clients has lead to the research direction of `fair' FL \cite{TowardsFairFL}.

\begin{algorithm}[h]
\caption{Federated Averaging \cite{FedAvgPaper} }
\textbf{input:} initial global model $\bm{x}_0$, learning rate schedule $\{ \eta_r \}$, local steps schedule $\{ K_r \}$  \\
\For{\rm{round} $r = 1 \ \textbf{to} \ R$}{
	select round clients $\mathcal{C}_r$ \\
	\For{\rm{client} $c \in \mathcal{C}_r$ \rm{in parallel}}{
		download global model $\bm{x}_r$ \\
		\For{\rm{local SGD step} $k = 1 \ \textbf{to} \ K_r$}{
			$\bm{x}^{c}_{r,k} \gets \bm{x}_r - \eta_r \nabla f(\bm{x}^{c}_k, \xi^{c}_{k})$
		}
		upload local model $\bm{x}^{c}_{r,K_r}$ to server
	}
	update global model $\bm{x}_{r+1} \gets \frac{1}{|C_r|} \sum_{c \in C_r} \bm{x}^{c}_{r,K_r}$
}
\end{algorithm}

The updates to clients models within the FedAvg process can be viewed from the perspective of communication rounds (as shown in most FL works and presented in Algorithm 1), but can also be reformulated in terms of a continuous sequence of SGD steps on each client, with updates periodically being replaced by averaging. Suppose we reindex the client models from $\bm{x}^{c}_{r,k}$ to $\bm{x}^{c}_{t}$ where $t$ is the \textit{global iteration}, $t \in \{1, \cdots, T \}$. Note that $\sum_r \{ K_r \} = T$. For a given FL client $i$ and local SGD step $t$ the update to the local model $\bm{x}^i_t$ can be given as:

\begin{equation}
\begin{aligned}
\bm{y}^{i}_{t+1} &= \bm{x}^i_t - \eta_t \nabla f(\bm{x}^{i}_t, \xi^{i}_{t}), \\
\bm{x}^{i}_{t+1} &= 
	\begin{cases}
		\sum_{c \in \mathcal{C}_t} p_c \bm{y}^c_{t+1} & \text{if} \  t \in \mathcal{I}, \\
		\bm{y}^i_{t+1} & \text{otherwise},
	\end{cases}
\end{aligned}
\end{equation}

\noindent where $\mathcal{I}$ is the set of indexes denoting the iterations at which model communication occurs (which will be equal to the cumulative sum of $\{ K_r \}$). This formulation states that clients not participating in the current round compute and then discard some local updates, which is not true in reality but makes analysis more amenable and is theoretically equivalent to FedAvg as presented in Algorithm 1. We define the average client model at any given iteration $t$ using (2) as: $\bar{\bm{x}}_t = \sum_{c = 1}^C p_c \bm{x}_c^t$.

\subsection{Runtime Model of FedAvg}
Inspecting Algorithm 1 shows that the nominal wall-clock time for each client $c$ to complete a communication round $r$ is:
\begin{equation}
W^c_r = \frac{|\bm{x}|}{D^c} + K_r \beta^c + \frac{|\bm{x}|}{U^c},
\end{equation}
\noindent where $|\bm{x}|$ is the size of the FL model (in megabits), $U^c$ and $D^c$ are the upload and download bandwidth of client $c$ in megabits per second, and $\beta^c$ is the per-minibatch computation time of client $c$. The nominal time to complete a round for client $c$ is therefore the sum of the download, local compute, and upload times. Furthermore, as FL clients are usually connected wirelessly at the network edge and geographically dispersed, we assume that $U^c$ and $D^c$ are independent of the total number of participating clients. For wireless connections, typically $D^c >> U^c$.

For a single round, the server must wait for the slowest client (straggler) to send its update. Therefore the time taken to complete a single round $W_r$ is: 
\begin{equation}
W_r = \underset{c \in \mathcal{C}_r}{\max} \left\lbrace { W^c_r } \right\rbrace.
\end{equation}
\noindent To simplify the FedAvg runtime model, we assume that all clients have the same upload bandwidth, download bandwidth, and per-minibatch compute time. That is, $U^c = U$, $D^c = C$, and $\beta^c = \beta$, $\forall \ c$. Using these simplifications, the total runtime $W$ for $R$ communication rounds of FedAvg are:
\begin{equation}
W = \sum_{r = 1}^R W_r = R \left( \frac{|\bm{x}|}{D} + \frac{|\bm{x}|}{U} \right) + \beta \sum_{r = 1}^R K_r.
\end{equation}
Previous works consider a fixed number of local steps during training: $K_r = K, \ \forall \ r$. There are extensive works showing that a larger $K$ can lead to an increased convergence rate of the global model \cite{FedAvgPaper}. However, large $K$ means that fewer communication rounds can be completed in a given timeframe. Previous works have shown that due to the low computational power of FL clients, the value of $\beta$ can dominate the per-round runtime \cite{TowardsEffSched}. Therefore by decaying $K_r$ during training, a balance between fast convergence and higher round-completion rate can be achieved, which is the primary focus of this work.

\subsection{Convergence Analysis}
We now present a convergence analysis of FedAvg using a decaying number of local steps $K_r$ and constant learning rate $\eta$. We make the following assumptions, which are typical of within the theoretical analysis within FL.
\\

\noindent \textbf{Assumption 1:} Client objective functions are $L$-smooth: 
\begin{equation*}
f_c(\bm{x}) \leq f_c(\bm{y}) + (\bm{x}-\bm{y})^{\top} \nabla f_c (\bm{y}) + \frac{L}{2}\| \bm{x} - \bm{y} \|^2.
\end{equation*}
As $F(\bm{x})$ is a convex combination of $f_c$, then it is also $L$-smooth.
\\

\noindent \textbf{Assumption 2:} Client objective functions are $\mu$-strongly convex: 
\begin{equation*}
f_c(\bm{x}) \geq f_c(\bm{y}) + (\bm{x}-\bm{y})^{\top} \nabla f_c (\bm{y}) + \frac{\mu}{2}\| \bm{x} - \bm{y} \|^2,
\end{equation*}
with minima $f^*_c = \min{f_c}$. As $F(\bm{x})$ is a convex combination of $f_c$, then it is also $\mu$-strongly convex.
\\

\noindent \textbf{Assumption 3:} For uniformly sampled data points $\xi^{c}_{k}$ on client $c$, the variance of stochastic gradients on $c$ are bounded by: 
\begin{equation*}
\E \| \nabla f_c(\bm{x}; \xi^{c}_{k}) - \nabla f_c(\bm{x}) \|^2 \leq \sigma_c^2.
\end{equation*}
\\

\noindent Due to analysing gradient descent on an $L$-smooth function, the magnitude of the gradient is naturally bounded by the distance between the first iterate $\bm{x}_1$ and the minimiser:
\begin{equation*}
\| \nabla F( \bm{x} )  \|^2 \leq L^2 \| \bm{x}_1 - \bm{x}^* \|^2 .
\end{equation*}
In our later analysis we define $G^2 = L^2 \| \bm{x}_1 - \bm{x}^* \|^2$ for convenience, i.e. the maximum norm of the gradient during training.
\\

\noindent As per \cite{ConvergenceFedAvgNonIID}, we quantify the extent of non-IID client data with:
\begin{equation*}
\Gamma = F^* - \sum_{c = 1}^{C}p_c f_c^*,
\end{equation*}
where $F^*$ is the minimum point of $F(\bm{x})$. $\Gamma \neq 0$ when the minimiser of the global objectives is not the same as the average minimiser of client objectives. $\Gamma = 0$ if the FL data is IID over the clients. 

Assumption 2 states that our analysis considers strongly-convex objectives. Although FL is typically used to train large DNNs (with nonconvex objectives), strongly-convex models are widely-used, for example in Support Vector Machines. Furthermore, the starte-of-the-art in analysing FedAvg's convergence behaviour lags behind the empirical developments, with contemporary anlyses also making the convex assumption \cite{ConvergenceFedAvgNonIID} \cite{FLOverWirelessNetworks}. The experimental evaluations in Section 4 consider one convex model (Sentiment 140) and three nonconvex DNNs. We leave it to future work to derive optimal $K$ schedules for nonconvex objectives. 
\\

\noindent \textbf{Theorem 1:} \textit{Let Assumptions 1-3 hold, and define $\kappa = \nicefrac{L}{\mu}$. The expected minimum gradient norm of FedAvg using a monotonically decreasing number of local SGD steps $K_r$ and fixed stepsize $\eta \leq \nicefrac{1}{4L}$ after $T$ total iterations is given by: }
\begin{multline}
\underset{t}{\min} \lbrace \E \left[ \| \nabla F(\bar{\bm{x}}_t) \|^2 \right] \rbrace \leq \frac{2 \kappa (\kappa F(\bar{\bm{x}}_0) - F^*)}{\eta T} \\ 
+ \eta \kappa L \left[ \sum^{C}_{c = 1} p^2_c \sigma^2_c + 6 L \Gamma + \left( 8 + \frac{4}{N} \right)  G^2 \frac{\sum_{r = 1}^R K_r^3}{\sum_{r = 1}^{R} K_r } \right]. 
\end{multline}
\textbf{Proof:} See Appendix A.2.
\\

\noindent The above theorem provides some useful insights into the convergence properties of FedAvg when using multiple local steps. Some of these are detailed below. \\

\noindent \textbf{Remark 1.1: relation to centralised SGD.} With a fixed learning rate and decreasing $K_r$, Theorem 1 shows that FedAvg converges with $\mathcal{O}(\nicefrac{1}{T}) + \mathcal{O}(\eta)$. This result reflects the classical result of centralised SGD with a fixed learning rate (albeit with different constants due to non-IID clients and $K_r > 1$). Previous works have shown (like in centralised SGD) the requirement for $\eta$ to be decayed to allow FedAvg to converge arbitrarily close to the global minima \cite{SCAFFOLD,ConvergenceFedAvgNonIID}. However in this work we are interested in the runtime and computational savings available when decaying $K_r$, so do not feel the need to prove the already-covered decaying $\eta$ result here. \\

\noindent \textbf{Remark 1.2: benefit of $\bm{K > 1}$.} When using a decreasing $\eta$, previous analyses have shown that $K > 1$ acts to reduce the variance introduced by client stochastic gradients (the $\sum_{c=1}^C p_c^2 \sigma_c^2$ term) \cite{SCAFFOLD,ConvergenceFedAvgNonIID}. Dividing (6) by $K_r$ (to achieve the convergence result in terms of total number of rounds) shows the same benefit in our analysis. We also observe empirically that $K_r > 1$ helps to reduce the variance of the global model updates even with a fixed $\eta$. Similarly a large  number of clients participating per round ($N$) helps to reduce the variance that is introduced by performing $K_r > 1$ steps. FedAvg deployments can therefore benefit more from sampling a larger number of clients per round $N$ when the number of local steps $K_r$ is large. \\

\noindent \textbf{Remark 1.3: drawback of $\bm{K > 1}$.} The second term of Theorem 1 shows that using $K_r > 1$ harms the convergence of FedAvg in terms of total number of iterations $T$. This is the case for all state-of-the-art analyses save for quadratic objectives \cite{IsLocalSGDBetter}. However, in FL we wish to minimise the number of communication rounds (due to the quantity of communicated data and impact of stragglers etc.) alongside the total number of iterations (both of which affect the runtime of FedAvg). \\

\noindent \textbf{Remark 1.4: real-world participation rates.} Our formulation of FedAvg our analysis assumes a constant participation rate, but in real-world FL the round participation rate varies \cite{FLAtScale}. Setting $K_r$ to a large value makes more progress in a round, but fewer clients will be able to complete the round in a given timeframe. This poses an interesting trade-off between $K_r$ and $N$, which could be a potential avenue for future research.

\begin{table*}[h]
\normalsize
\centering
\caption{Datasets and models used in the experimental evaluation (DNN = Deep Neural Network, CNN = Convolutional Neural Network, GRU = Gated Recurrent Network). $K_0$ and $\eta_0$ are the initial number of local steps and initial learning rate used.}
\begin{tabular}{ c | c c c c c c c c c}
 	\multirow{2}{*}{Task} & \multirow{2}{*}{Type} & \multirow{2}{*}{Classes} & \multirow{2}{*}{Model} & Model & \multirow{2}{*}{Total Clients} & Clients & Samples & \multirow{2}{*}{$K_0$} & \multirow{2}{*}{$\eta_0$} \\
	& & & & Size (Mb) & & per Round & per Client \\ 
	\hline
	Sent140 	& Sentiment analysis 	& 2    & Linear & 0.32 & 21876 & 50  & 15   & 60 & 3.0 \\
	FEMNIST 	& Image classification 	& 62   & DNN 	& 6.71 & 3000  & 60  & 170  & 80 & 0.3  \\
	CIFAR100 	& Image classification 	& 100  & CNN 	& 40.0 & 500   & 25  & 100  & 50 & 0.01 \\
	Shakespeare & Character prediction 	& 79   & GRU 	& 5.21 & 660   & 10  & 5573 & 80 & 0.1  \\
\end{tabular}
\end{table*}

\subsection{Optimal Values of $K_r$ and $\eta_r$}
FedAvg is an iterative algorithm with each round starting from a new global model. Therefore, each iteration can be viewed as restarting the algorithm, using model $\bm{x}_{t_0}, \forall t_0 \in \mathcal{I}$. Using this formulation, we can indepedently derive what the optimal fixed value of $K$ would be at the start of any communication round during training. As training progresses and new rounds of training are completed, this value of $K$ can therefore vary. When using a fixed number of local steps $K_r = K$ and communication rounds $R$, the total number of FedAvg iterations is $T = KR$. Substituting this into (5) gives the total runtime $W$ of $T$ iterations of FedAvg:
\begin{equation}
W = \frac{T}{K} \left[ \frac{|\bm{x}|}{D} + \frac{|\bm{x}|}{U} + \beta K \right].
\end{equation}
Setting $K_r = K$ and substituting (7) into Theorem 1 gives us the convergence of $t$ rounds of FedAvg for fixed $K$ and $\eta$ in terms of the runtime, starting from an arbitrary round in the training process $\bm{x}_{t_0}, \forall t \in \mathcal{I}$, rather than the number of iterations:
\begin{multline}
\underset{t > t_0}{\min} \lbrace \E \left[ \| \nabla F(\bar{\bm{x}}_t) \|^2 \right] \rbrace \\
\leq \frac{2 \kappa (\kappa F(\bar{\bm{x}}_{t_0}) - F^*)}{\eta W K} \left[ \frac{|\bm{x}|}{D} + \frac{|\bm{x}|}{U} + \beta K \right] \\ 
+ \eta \kappa L \left[ \sum^{C}_{c = 1} p^2_c \sigma^2_c + 6 L \Gamma + \left( 8 + \frac{4}{N} \right) G^2 K^2 \right]. 
\end{multline}
As (8) gives the convergence of $t$ rounds of FedAvg, starting from arbitrary point $\bm{x}_{t_0}$, with a fixed $K$ and $\eta$. If the round index is instead substituted with a time index (with $x_w$ corresponding to the value of $x_t$ at time $w$), (8) can be used to determine what the optimal fixed valued of $K$ looking forward would be for any point in time during the training process, $K_w^*$. 
\\

\noindent \textbf{Theorem 2:} \textit{Let Assumptions 1-3 hold and define $\kappa = \nicefrac{L}{\mu}$. For fixed $\eta \leq \nicefrac{1}{4L}$, the optimal number of local SGD steps $K$ to minimise (8) is given by:}
\begin{equation}
K_w^* = \sqrt[3]{ \frac{ \left( \kappa F(\bar{\bm{x}}_{t_0}) - F^* \right)}{8 \eta^2 L \left( 1 + \nicefrac{1}{2N}\right)} \frac{\left( \nicefrac{|\bm{x}|}{D} + \nicefrac{|\bm{x}|}{U} \right) }{ W } }.
\end{equation}
\textbf{Proof:} See Appendix A.3.
\\

\noindent Theorem 2 shows that $K_w^*$ decreases at least as fast as $\mathcal{O}(\nicefrac{1}{\sqrt[3]{W}})$, motivating the principal of decreasing the number of local steps during FedAvg. The decreasing value of the global model objective $F(\bar{\bm{x}}_{t_0})$ also influences $K_w^*$. The implications of Theorem 2 are discussed in the following Remarks. \\

\noindent \textbf{Remark 2.1: relation to other works.} Wang and Joshi \cite{AdaptiveCommStrats} investigated variable communication intervals for the Periodic Averaging SGD (PASGD) algorithm in the datacentre, and found that the optimal interval decreased with $\mathcal{O}(\nicefrac{1}{\sqrt[2]{W}})$. $K^*_w$ decreases slower in FedAvg due to the looser bound on client divergence between averaging (scaling with $K^2$ rather than $K$). \\

\noindent \textbf{Remark 2.2: dependence on client participation rate.} As the number of clients participating per round ($N$) increases, $K^*_w$ increases. This is because a higher number of participating clients decreases the variance in model updates (which is especially significant considering the non-IID client data). \\

\noindent \textbf{Remark 2.3: reformulation using communication rounds.} In FL, it is typically assumed that the local computation time is dominated by the communication time due to the low-bandwidth connections to the coordinating server. If we consider the case where ($\nicefrac{|\bm{x}|}{D} + \nicefrac{|\bm{x}|}{U} >> \beta K$), then $W \approx R \left( \nicefrac{|\bm{x}|}{D} + \nicefrac{|\bm{x}|}{U} \right)$. This means:
\begin{align}
K_r^* 	&= 		\sqrt[3]{ \frac{ \kappa F(\bar{\bm{x}}_{t_0}) - F^* }{8 \eta^2 L \left( 1 + \nicefrac{1}{2N}\right) }\frac{1}{R} } \nonumber \\
		&\leq 	\sqrt[3]{ \frac{ \kappa F(\bar{\bm{x}}_0) - F^* }{8 \eta^2 L \left( 1 + \nicefrac{1}{2N}\right) }\frac{1}{R} },	
\end{align}
where the inequality comes from the fact that $F(\bm{x}_t) \leq F(\bm{x}_0)$ given Assumption 1 and an appropraitely chosen stepsize $\eta$. $K^*_r$ is not dependent on the local computation time, only the total number of rounds $R$. Using (10) as a decay scheme produces a fairly aggressive decay rate, and is tested experimentally in Section 4 using a variety of model types (which have different communication and computation times). \\

\noindent A similar approach can be taken to find the optimal value of $\eta^*_r$ at each communication round. Although the focus of this paper is on decaying $K$ to improve the convergence speed of FL, we compare it to the effect of decaying $\eta$ as well as constant $\eta$ and $K$. 
\\

\noindent \textbf{Corollary 2.1: } \textit{Let Assumptions 1-3 hold and define $\kappa = \nicefrac{L}{\mu}$. Given stepsizes $\eta_r \leq \nicefrac{1}{4L}$, the optimal value of $\eta$ at any point in time during training to minimise (8) is given by:}
\begin{multline}
\eta_w^* = \sqrt{ \frac{ 2( \kappa F(\bar{\bm{x}}_{t_0}) - F^*)}{LZ}  \frac{\left( \nicefrac{|\bm{x}|}{D} + \nicefrac{|\bm{x}|}{U} + \beta K \right) }{ W } }, \\
\text{where} \ Z = \sum_{c=1}^C p_c^2 \sigma_c^2 + 6L\Gamma + (8 + \nicefrac{4}{N})G^2 K^2.
\end{multline}
\textbf{Proof:} See Appendix A.4.
\\

\noindent Corollary 2.1 shows that the optimal value of $\eta$ decreases with $\mathcal{O}(\nicefrac{1}{\sqrt{W}})$. Several insights from Corollary 2.1 are given below. \\

\noindent \textbf{Remark 2.1.1: impact of round time.} (11) shows that $\eta^*_w$ is directly affected by the per-round time: as any of the upload, download or computation time increases, $\eta^*_w$ increases. This is because less progress is made over time (due to longer rounds) so a larger $\eta^*+W$ compensates by making more progress per SGD step. \\

\noindent \textbf{Remark 2.1.2: dependence on client participation rate.} Similar to $K_w^*$, a larger number of clients participating per round ($N$) allows for a smaller $\eta_w^*$ by reducing variance due to client-drift. Larger $K$ in (11) also allows for smaller $\eta$ as more progress is made per round. \\

\noindent \textbf{Remark 2.1.3: reformulation using communication rounds.} Making the same assumption as in (10) ($\nicefrac{|\bm{x}|}{D} + \nicefrac{|\bm{x}|}{U} >> \beta K$) gives a decay schedule for $\eta_r$ in terms of the number of communication rounds:
\begin{align}
\eta_r^* 	&= 		\sqrt{\frac{2(\kappa F(\bm{x}_{t_0}) - F^*)}{LZ} \frac{1}{R}} \nonumber \\
			&\leq 	\sqrt{\frac{2(\kappa F(\bm{x}_0) - F^*)}{LZ} \frac{1}{R}},
\end{align}
where $Z$ is defined in (11), and the inequality again comes from using $F(\bm{x}_t) \leq F(\bm{x}_0)$. This decay schedule is also tested empirically in Section 4.

\subsection{Schedules Based on Training Progress}
In practice the values of $\kappa$, $F^*$, and $L$ are difficult or impossible to evaluate due to complex nonlinear models (i.e. DNNs) and data privacy in FL. Therefore, appropriate values of $K$ and $\eta$ are chosen via grid-search or some other method (such as Bayesian Optimisation). Denote $K_0$ as a `good' value of $K$ at $W = 0$ (found via grid search), and $K_r$ as the value of $K$ to be used for round $r$. Each successive round of FedAvg can be considered as a new optimisation procedure with starting model $\bar{\bm{x}}_r$. If we make the further assumption that $F^* = 0$, substituting these two sets of values into (9) and dividing gives us $K_r$ in terms of $K_0$:
\begin{equation}
K_r^* = \left\lceil \sqrt[3]{\frac{F(\bar{\bm{x}}_r)}{F(\bar{\bm{x}}_0)}} \ K_0 \right\rceil.
\end{equation}
A similar process can be applied to find $\eta_r$ in terms of $\eta_0$:
\begin{equation}
\eta_r^* = \sqrt[2]{\frac{F(\bar{\bm{x}}_r)}{F(\bar{\bm{x}}_0)}} \ \eta_0.
\end{equation}
$F(\bar{\bm{x}}_r)$ is the training loss of the global model at the start of round $r$. Practically, an estimate of $F(\bar{\bm{x}}_r)$ can be obtained by requiring clients $c \in \mathcal{C}_r$ to send their training loss after the first step of local SGD to the server each round: $f_c(\bar{\bm{x}}_r, \xi_{c,0})$, where $\E \left[ f_c(\bar{\bm{x}}_r, \xi_{c,0}) \right] = F(\bar{\bm{x}}_r)$. This is only a single floating-point value that does not require any extra computation and negligibly increases the per-round communication costs.

Due to only a small fraction of the non-IID clients being sampled per round, the per-round variance of $\frac{1}{N} \sum_{c \in \mathcal{C}_r} f_c(\bar{\bm{x}}_r, \xi_{c,0})$ can be very high. Therefore, we propose a simple rolling-average estimate using window size $s$: 
\begin{equation}
F(\bar{\bm{x}}_r) \approx \frac{1}{sN} \sum_{i = r-s}^r \sum_{c \in \mathcal{C}_i} f_c(\bar{\bm{x}}_i, \xi_{c,0}).
\end{equation}
Our experiments in Section 4 use a window size $s = 100$, where our experiments run for at least $R = 10,000$ communication rounds. For the first $s$ rounds when (15) cannot be computed, we keep $K_r = K_0$.  

When using a fixed value of $K$, Theorem 1 shows that the minimum gradient norm converges with $\mathcal{O}(\nicefrac{1}{T}) + \mathcal{O}(\eta K^2)$. As noted earlier, this result is analogous to the classical result of dSGD using a fixed learning rate. In the datacentre, the practical heuristic of decaying the learning rate $\eta$ when the validation error plateaus is commonly used to allow the model to reach a lower validation error. We can therefore use a similar strategy for FedAvg: once the validation error plateaus we decay either $K$ or $\eta$. We investigate this heuristic alongside the decay schedules presented above in Section 4.

\begin{figure*}[h]
\includegraphics[width=\textwidth]{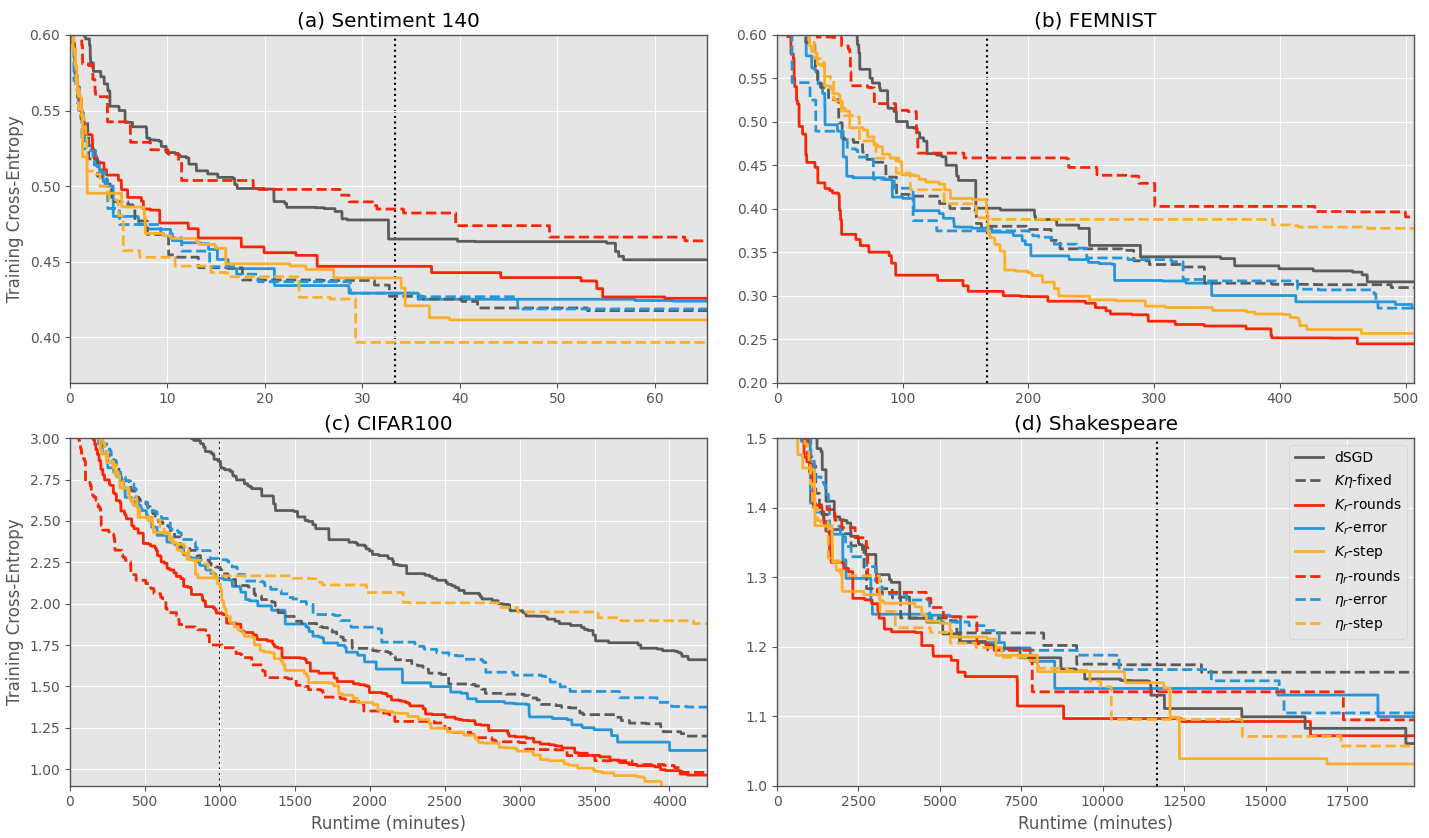}
\caption{Cumulative lowest training cross-entropy error over time of FedAvg using different schedules for $K_r$ and $\eta_r$. Curves show mean over 5 random trials. Vertical line shows the communication round where the validation error plateaus.}
\end{figure*}

\section{Experimental Evaluation}
In this section, we present the results of simulations comparing the three decaying-$K$ schemes proposed in Section 3 to evaluate their benefits in terms of runtime, communicated data and computational cost on four benchmark FL datasets. Code to reproduce the experiments is available from: \textit{github.com/JedMills/Faster-FL}.

\subsection{Datasets and Models}
To show the broad applicability of our approach, we conduct experiments on 4 benchmark FL learning tasks from 3 Machine Learning domains (sentiment analysis, image classification, sequence prediction) using 4 different model types (simple linear, DNN, Convolutional, Recurrent).
\\

\noindent \textbf{Sentiment 140:} a sentiment analysis task of Tweets from a large number of Twitter users \cite{LEAF}. We limited this dataset to users with $\geq 10$ samples, leaving 22k total clients, with 336k training and 95k validation samples, with an average of 15 training samples per client. We generated a normalised bag-of-words vector of size 5k for each sample using the 5k most frequent tokens in the dataset. We train a binary linear classifier (i.e. a convex model) using 50 clients per round ($0.2\%$ of all clients) and a batch size of $8$.
\\

\noindent \textbf{FEMNIST:} an image classification task of ($28 \times 28$) pixel greyscale (flattened) images of handwritten letters and numbers from 62 classes, grouped by the writer of the symbol \cite{LEAF}. We used 3k total clients, with a total of 501k training and 129k validation samples, with an average of 170 training samples per client. We sample 60 clients per round ($2\%$ of all clients) with a batch size of 32. We train a DNN comprising a 200-unit ReLU Fully-Connected layer (FC), a second 200-unit ReLU FC layer, and a softmax output layer.
\\

\noindent \textbf{CIFAR100:} an image classification task of ($32 \times 32$) pixel RGB images of objects from 100 classes. We use the non-IID partition first proposed in \cite{AdaptiveFedOpt}, which splits the images into 500 clients based on the class labels. There are 50k training and 10k validation samples in the dataset, with each client possessing 100 samples. We select 25 clients per round ($5\%$ of all clients). We train a Convolutional Neural Network (CNN) consisting of two ($3 \times 3$) ReLU convolutional + ($2 \times 2$) Max-Pooling blocks, a 512-unit ReLU FC layer, and a softmax output layer. As per other FL works \cite{AdaptiveFedOpt,Mime} we apply random preprocessing composed of a random horizontal flip and crop of the ($28 \times 28$) pixel sub-image to improve generalisation.
\\

\noindent \textbf{Shakespeare:} a next-character prediction task using the complete plays of William Shakespeare \cite{LEAF}. The lines from all plays are partitioned by the speaking part in each play, and clients with $\leq 2$ lines are discarded, leaving 660 total clients. Using a sequence length of 80, there are 3.7m training and 357k validation samples, with an average of 5573 training samples per client. We sample 10 clients per round ($1.5\%$ of all clients) with a batch size of $32$. We train a Gated Recurrent Unit (GRU) DNN comprising a $79 \to 8$ embedding, two stacked GRUs of 128 units, and a softmax output layer.
\\

\begin{figure*}[h]
\includegraphics[width=\textwidth]{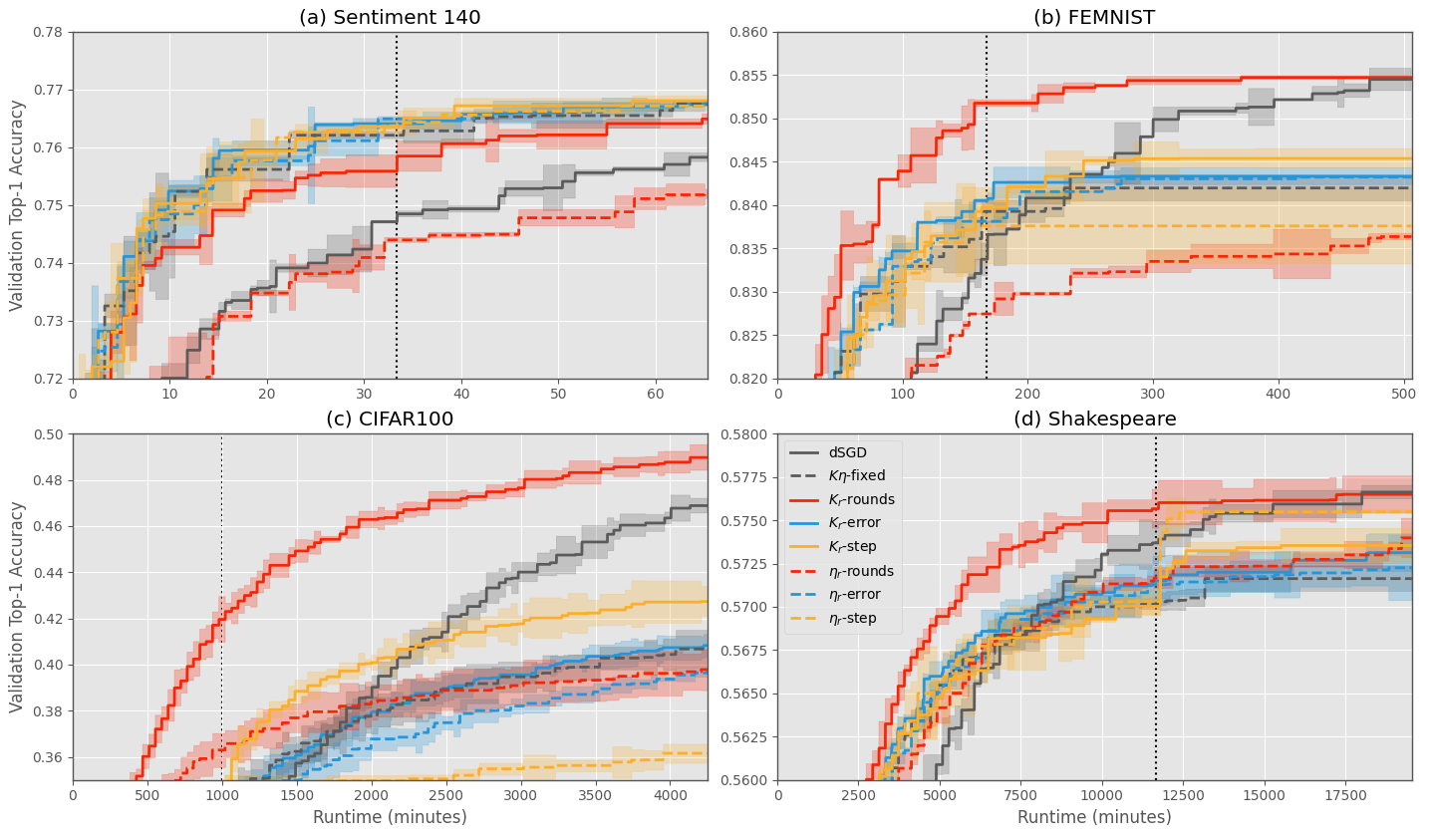}
\caption{Cumulative highest validation top-1 accuracy over time of FedAvg using different schedules for $K_r$ and $\eta_r$. Curves show mean and shaded regions show $95\%$ confidence intervals of the mean over 5 random trials.}
\end{figure*}

\subsection{Simulating Communication and Computation}
The convergence of FedAvg for the learning tasks was simulated using Pytorch on GPU-equipped workstations. However, real-world FL runs distributed training on low-powered edge clients (such as smartphones and IoT devices). These clients exhibit much lower computational power and lower bandwidth to the coordinating server compared to datacentre nodes. 

To realistically simulate real-world FedAvg, we use the runtime model presented in Section 3.2 and Equation (5). We assume that each client has a download bandwidth of $D = 20 \ \text{Mbps}$ and an upload bandwidth of $U = 5 \ \text{Mbps}$. These are typical values for wireless devices connected via 4G LTE in the United Kingdom \cite{OpenSignal}. To determine the runtime of a minibatch of SGD on a typical low-powered edge device ($\beta$), we ran 100 steps of SGD for each learning task on a Raspberry Pi 3B+ with the following configuration: 
\begin{itemize}
\item{1.4GHz 64-bit quad-core Cortex-A53 processor.}
\item{1GB LPDDR2 SDRAM.}
\item{Ubuntu Server 22.04.1.}
\item{PyTorch 1.8.2.}
\end{itemize}
Table 2 presents the values of $\beta$ recorded.

{\renewcommand{\arraystretch}{1.1}%
\begin{table}[h]
\normalsize
\centering
\caption{Mean and standard deviation of runtimes for a minibatch of SGD for each learning task using a Raspberry Pi 3B+.}
\begin{tabular}{ c | c c }
 	Task & Mean ($s$) & Std ($s$) \\
	\hline
	Sent140 	& $5.2\times 10^{-3}$ & $2.1 \times 10^{-4}$  \\
	FEMNIST 	& 0.017  & $5.1 \times 10^{-4}$  \\
	CIFAR100 	& 0.31   & $1.7 \times 10^{-2}$  \\
	Shakespeare & 1.5    & $8.5 \times 10^{-2}$  \\
\end{tabular}
\end{table}
}

As shown in Table 2, there is a large difference in the minibatch runtimes between the tasks. This is due to the relative computational costs of the models used: the Sent140 task uses a simple linear model, whereas the Shakespeare GRU model requires a far larger number of matrix multiplications for a single forward-backward pass.

\subsection{$K_r$ and $\eta_r$ Decay Schedules}
For each learning task, we ran FedAvg for 10k communication rounds using fixed $K_r = K_0$ and $\eta_r = \eta_0$ (henceforth `$K\eta$-fixed'). The number of rounds reflects typical real-world deployments (which are on the order of thousands of rounds) \cite{FLAtScale}. We selected $K_0$ and $\eta_0$ via grid-search such that the validation error for each task could plateau within the 10k rounds, and present the values in Table 1. We also ran dSGD (FedAvg with $K_r = 1$) to show the runtime benefit of using $K > 1$ local steps.

We then ran FedAvg using the three schedules for $K_r$ and the three schedules for $\eta_r$ as discussed in Section 3.4 and 3.5. Table 3 shows the different decay schedules tested and the name we denote each one by in Section 4.4. We also tested jointly decaying $K_r$ \textit{and} $\eta_r$ during training. However decaying either $K_r$ or $\eta_r$ decreases the amount of progress that is made during each training round as the global model changes less. We found empirically that decaying both lead to training progress slowing too rapidly, so have not included the results in Section 4.4.

{\renewcommand{\arraystretch}{1.3}%
\begin{table}[h]
\normalsize
\centering
\caption{Values of $K_r$ and $\eta_r$ for given communication round $r$ as tested in the experimental evaluation.}
\begin{tabular}{ c | c c }
 	Schedule & $K_r$ & $\eta_r$ \\
	\hline
	dSGD 			& 1 & $\eta_0$ \\
	$K\eta$-fixed 	& $K_0$ & $\eta_0$ \\
	\hline 
	$K_r$-rounds (10) 	& $\lceil \sqrt[3]{1/r} \ K_0 \rceil$ & $\eta_0$ \\
	$K_r$-error (13)	& $\lceil \sqrt[3]{\nicefrac{F_r}{F_0}} \ K_0 \rceil$ & $\eta_0$ \\
	$K_r$-step 		& $\nicefrac{K_0}{10}$ if converged & $\eta_0$ \\
	\hline
	$\eta_r$-rounds (12) & $K_0$ & $\sqrt[2]{1/r} \ \eta_0$ \\
	$\eta_r$-error (14)	& $K_0$ & $\sqrt[2]{\nicefrac{F_r}{F_0}} \ \eta_0 $\\
	$\eta_r$-step 	& $K_0$ & $\nicefrac{\eta_0}{10}$ if converged \\
\end{tabular}
\end{table}
}

{\renewcommand{\arraystretch}{1.1}%
\begin{table}[h]
\centering
\caption{Total SGD steps performed during training for each $K$-decay schedule relative to $K\eta$-fixed for different learning tasks.}
\begin{tabular}{ c c | c }
 	\multirow{2}{*}{Task }& \multirow{2}{*}{Schedule} & Relative \\
 													& & SGD Steps \\
	\hline
	\multirow{3}{*}{Sentiment 140} 	& $K_r$-rounds  & 0.21 \\
									& $K_r$-error   & 0.99 \\
									& $K_r$-step	& 0.68 \\
	\hline
	\multirow{3}{*}{FEMNIST} 		& $K_r$-rounds  & 0.11 \\
									& $K_r$-error   & 0.80 \\
									& $K_r$-step	& 0.44 \\
	\hline
	\multirow{3}{*}{CIFAR100} 		& $K_r$-rounds  & 0.090 \\
									& $K_r$-error   & 0.57  \\
									& $K_r$-step	& 0.40  \\
	\hline
	\multirow{3}{*}{Shakespeare} 	& $K_r$-rounds  & 0.74 \\
									& $K_r$-error   & 0.99 \\
									& $K_r$-step	& 0.96 \\
\end{tabular}
\end{table}
}

\subsection{Results}
Figure 1 shows the minimum cumulative training error achieved by FedAvg for the different $K_r$ and $\eta_r$ schedules (as shown in Table 3). Confidence intervals for Figure 1 were omitted for clarity due to the larger number of curves. For all tasks other than Shakespeare, FedAvg with $K\eta$-fixed (solid grey curve) increases the convergence rate compared to dSGD (dashed grey curve). For Shakespeare (Figure 1 (d)), $K\eta$-fixed improved the initial convergence rate but was overtaken by dSGD at approximately 2500 minutes. This is likely because of the very high computation time for Shakespeare (see Table 2) relative to the other datasets (due to the very high computational cost of the GRU model). 

For Sentiment 140 (Figure 1 (a)) and Shakespeare (Figure 1 (d)), decaying either $K_r$ or $\eta_r$ during training lead to smaller improvements in the training error that was achieved. However, for FEMNIST and CIFAR100, the $K_r$-rounds scheme lead to lower training error compared to $K\eta$-fixed. For CIFAR100, an improvement was also seen with $\eta_r$-rounds. Both FEMNIST and CIFAR100 are image classification tasks, so it be may the case that decaying $K_r$ or $\eta_r$ during training is beneficial for computer vision tasks, which could be investigated further in future works.

Figure 2 shows the impact on validation accuracy for the tested decay schedules. The $K\eta$-fixed schedule shows faster initial convergence for all tasks, but it is overtaken by dSGD in the later stages of training. For FEMNIST, CIFAR100 and Shakespeare, the aggressive $K_r$-rounds and $K_r$-step schemes improved the convergence rate compared to dSGD, with very significant improvement for CIFAR100. A marked increase in convergence rate can be seen in Figure 1 (c) at 1000 minutes when $K_r$-step is decayed. 

In all tasks, all $K$-decay schemes were able to match or improve the validation accuracy that $K\eta$-fixed achieved whilst performing (often substantially) fewer total steps of SGD within a given runtime. Table 4 shows the total SGD steps performed by the $K$-decay schemes relative to the total steps performed by $K\eta$-fixed over the 10k communication rounds (all the $\eta$-decay schemes perform the same amount of computation as $K\eta$-fixed). The fact that $K$-decay schemes can outperform $K\eta$fixed with lower total computation indicates that much of the extra computation performed by FedAvg is wasted when considering validation performance. CIFAR100 using $K_r$-rounds for example achieved over $18\%$ higher validation accuracy compared to $K\eta$-fixed whilst performing less than $10\%$ of the total steps of SGD. Similarly, $K_r$-step achieved the same validation accuracy as $K\eta$-fixed whilst performing only $68\%$ of the total SGD steps.

\section{Conclusion}
The popular Federated Averaging (FedAvg) algorithm is used within the Federated Learning (FL) paradigm to improve the convergence rate of an FL model by performing several steps of SGD ($K$) locally during  each training round. In this paper, we analysed FedAvg to examine the runtime benefit of decreasing ($K$) during training. We set up a runtime model of FedAvg and used this to determine the optimal value of $K$ (and learning rate $\eta$) at any point during training under different assumptions, leading to three practical schedules for decaying $K$ as training progresses. Simulated experiments using realistic values for communication-time and computation-time on 4 benchmark FL datasets from 3 learning domains showed that decaying $K$ during training can lead to improved training error and validation accuracy within a given timeframe, in some cases whilst performing over $10\times$ less computation compared to fixed $K$.

\bibliographystyle{ieee}
\bibliography{k_decay_biblio}

\vfill
\clearpage

\appendices

\section{Proof of Theorems}

\subsection{Key Lemmas}
Previously, Li \textit{et al.} \cite{ConvergenceFedAvgNonIID} analysed the per-iteration convergence of FedAvg for $\mu$-strongly convex functions when using a decreasing stepsize. Their result was the first to prove convergence for non-IID clients with partial participation. We make assumptions that are at least as strong as Li \textit{et al.}, so can use their intermediary result bounding the distance to the global minimiser when using partial client participation:
\\

\noindent \textbf{Lemma 1:} \textit{Given Assumptions 1 - 3, the expected distance between average client model $\bar{\bm{x}}_t$ and the global minimiser $\bm{x}^*$ is upper-bounded by:}
\begin{multline}
\E \left [ \| \bar{\bm{x}}_{t+1} - \bm{x}^* \|^2 \right] \leq (1 - \eta_t \mu) \E \left[ \| \bar{\bm{x}}_t - \bm{x}^2 \|^2 \right]  \\ 
+ \eta_t^2 \Big( \sum_{c = 1}^{C} p_c^2 \sigma_c^2 + 6L\Gamma \\
 + 8(K_t - 1)^2 G^2 + \frac{C - N}{N - 1} \frac{4}{N} K_t^2 G^2 \Big).
\end{multline}
\textbf{Proof:} See Appendix B.3 of \cite{ConvergenceFedAvgNonIID}.
\\

\noindent \textbf{Lemma 2:} \textit{Given Assumptions 1 - 3, the sum of expected gradient norms over $T$ iterations of the average client model $\bar{\bm{x}}_t$ is upper-bounded by:}
\begin{multline}
\sum_{t=1}^T \eta_t \E \left[ \| \nabla F(\bar{\bm{x}}_t) \|^2 \right] \leq 2 \kappa (\kappa F(\bar{\bm{x}}_0) - F^*) \\ 
+ \kappa L \Big( \sum_{c = 1}^{C} p_c^2 \sigma_c^2 + 6L\Gamma \\
 + 8(K_t - 1)^2 G^2 + \frac{4}{N} K_t^2 G^2 \Big) \sum_{t=1}^T \eta_t^2.
\end{multline}
\textbf{Proof :} Rearranging Lemma 1 and then defining for notational convenience
$$
D = \left( \sum_{c = 1}^{C} p_c^2 \sigma_c^2 + 6L\Gamma  + 8(K_t - 2)^2 G^2 + \frac{C - N}{N - 1} \frac{4}{N} K_t^2 G^2 \right),
$$ the recursive definition can be written as:
\begin{multline}
\eta_t \mu \E \left [\| \bbmxt - \bm{x}^* \|^2 \right] \leq \E \left[ \| \bbmxt - \bm{x}^* \|^2 \right] \\
- \E \left[ \| \bar{\bm{x}}_{t+1} - \bm{x}^* \|^2 \right] + \eta_t^2 D.
\end{multline}
Using Assumption 1 ($L$-smoothness), we have:
\begin{multline}
\frac{\eta_t \mu}{L^2} \E \left[ \| \nabla F(\bbmxt) \|^2 \right] \leq \E \left[ \| \bbmxt - \bm{x}^* \|^2 \right] \\ 
- \E \left[ \| \bar{\bm{x}}_{t+1} - \bm{x}^* \|^2 \right] + \eta_t^2 D.
\end{multline}
Summing up the $T$ iterations and telescoping the distance terms gives:
\begin{multline}
\frac{\mu}{L^2} \sum_{t=1}^{T} \eta_t^2 \E \left[ \| \nabla F(\bbmxt) \|^2 \right] \leq \E \left[ \| \bar{\bm{x}}_0 - \bm{x}^* \|^2 \right] \\
- \E \left[ \| \bar{\bm{x}}_{T} - \bm{x}^* \|^2 \right] + D \sum_{t=1}^{T} \eta_t^2.
\end{multline}
Using Assumption 1 ($L$-smoothness) and Assumption 2 ($\mu$-strong convexity) to bound the distance terms now gives:
\begin{multline}
\frac{\mu}{L^2} \sum_{t=1}^{T} \eta_t^2 \E \left[ \| \nabla F(\bbmxt) \|^2 \right] \leq \frac{2}{\mu} \left[ F(\bar{\bm{x}}_0) - F^* \right] \\ - \frac{2}{L} \left[ F(\bar{\bm{x}}_T) - F^* \right] + D \sum_{t=1}^{T} \eta_t^2,
\end{multline}
which can be simplified by noting that $\mu \leq L$, so that:
\begin{multline}
\frac{\mu}{L^2} \sum_{t=1}^{T} \eta_t^2 \E \left[ \| \nabla F(\bbmxt) \|^2 \right] \leq \frac{2}{\mu} F(\bar{\bm{x}}_0) - \frac{2}{L} F(\bar{\bm{x}}_T) \\ + D \sum_{t=1}^{T} \eta_t^2.
\end{multline}
Multiplying both sides of the inequality by $L^2/\mu$, using the lower-bound $F^* \leq F(\bar{\bm{x}}_T)$, the definition $\kappa = L/\mu$, and the fact that $\frac{C - N}{N - 1} \leq 1$ completes the proof.

\subsection{Proof of Theorem 1}
The bound on gradient norms given in Lemma 2 uses the global index $t$ that denotes the global SGD step that each client evaluates (irrespective of communication rounds). However, FedAvg clients participate in communication rounds. The values of $\eta_t$ and $K_t$ are therefore fixed for each communication round. To account for this, Lemma 2 can be reindexed using the given communication round $r$ and local step $k$: $t = I + k$, where where $I = \sum_{i=1}^r K_i$. The total number of communication rounds is $R$, therefore $T = \sum_{r=1}^R K_r$.  Using this to reindex Lemma 2:
\begin{align*}
\sum_{r=1}^R \eta_r \sum_{k=1}^{K_r} & \E \left[ \| \nabla F(\bar{\bm{x}}_{I+k}) \|^2 \right] \\
&\leq 2 \kappa (\kappa F(\bar{\bm{x}}_0) - F^*) \\ 
& \quad + \kappa L \Big( \sum_{c = 1}^{C} p_c^2 \sigma_c^2 + 6L\Gamma + 8(K_r - 1)^2 G^2 \\
& \quad + \frac{4}{N} K_r^2 G^2 \Big) \sum_{r=1}^R \eta_r^2 K_r \\[6pt]
& \leq 2 \kappa \left( F(\bar{\bm{x}}_0) - F^* \right) \\
& \quad + \kappa L \left( \sum_{n=1}^N p_n^2\sigma_n^2 + 6L\Gamma \right) \sum_{r=1}^{R} \eta_r^2 K_r \\
& \quad + 16 \kappa L G^2 \sum_{r=1}^{R} \eta_r^2 K_r^3.
\end{align*}
Diving both sides through by $\sum_{r=1} \eta_r K_r$: 
\begin{align*}
& \frac{ \sum_{r=1}^R \eta_r \sum_{k=1}^{K_r} \E \left[ \| \nabla F(\bar{\bm{x}}_{I+k}) \|^2 \right] }{ \sum_{r=1}^R \eta_r K_r } \\
& \quad \leq \frac{2 \kappa \left( F(\bar{\bm{x}}_0) - F^* \right)}{\sum_{r=1}^R \eta_r K_r} \\
& \qquad + \kappa L \left( \sum_{n=1}^N p_n^2\sigma_n^2 + 6L\Gamma \right) \frac{\sum_{r=1}^{R} \eta_r^2 K_r}{\sum_{r=1}^{R} \eta_r K_r} \\
& \qquad + 16 \kappa L G^2 \frac{\sum_{r=1}^{R} \eta_r^2 K_r^3}{\sum_{r=1}^{R} \eta_r K_r}.
\end{align*} 
Using a fixed $\eta_r = \eta \leq 1/4L$, then the above inequality can be simplified as:
\begin{align*}
& \frac{ \sum_{r=1}^R \sum_{k=1}^{K_r} \E \left[ \| \nabla F(\bar{\bm{x}}_{I+k}) \|^2 \right] }{ \sum_{r=1}^R K_r } \\
& \quad \leq \frac{2 \kappa \left( F(\bar{\bm{x}}_0) - F^* \right)}{\eta \sum_{r=1}^R K_r} \\
& \qquad + \eta \kappa L \left( \sum_{n=1}^N p_n^2\sigma_n^2 + 6L\Gamma \right) \\
& \qquad + 16 \eta \kappa L G^2 \frac{\sum_{r=1}^{R} K_r^3}{\sum_{r=1}^{R} K_r}.
\end{align*} 
Reindexing using the fact that $\sum_{r=1}^R K_r = T$ gives:
\begin{align*}
& \frac{ \sum_{t=1}^T \E \left[ \| \nabla F(\bar{\bm{x}}_{t}) \|^2 \right] }{T} \\
& \quad \leq \frac{2 \kappa \left( F(\bar{\bm{x}}_0) - F^* \right)}{\eta T} \\
& \qquad + \eta \kappa L \left[ \sum_{n=1}^N p_n^2\sigma_n^2 + 6L\Gamma + 16 G^2 \frac{\sum_{r=1}^{R} K_r^3}{\sum_{r=1}^{R} K_r} \right]. \\
\end{align*}
Using $\min \lbrace \E \left[ \| \nabla F(\bar{\bm{x}}_t) \|^2 \right] \rbrace \leq \E \left[ \| \nabla F(\bar{\bm{x}}_{t}) \|^2 \right]$ then completes the proof.

\subsection{Proof of Theorem 2}
We start from the bound on gradient norms using a constant $K_w$ (within a communication round) and $\eta$ (8):
\begin{multline}
\underset{t > t_0}{\min} \lbrace \E \left[ \| \nabla F(\bar{\bm{x}}_{t_0}) \|^2 \right] \rbrace \\
\leq \frac{2 \kappa (\kappa F(\bar{\bm{x}}_{t_0}) - F^*)}{\eta W K_w} \left[ \frac{|\bm{x}|}{D} + \frac{|\bm{x}|}{U} + \beta K_w \right] \\ 
+ \eta \kappa L \left[ \sum^{C}_{c = 1} p^2_c \sigma^2_c + 6 L \Gamma + \left( 8 + \frac{4}{N} \right) G^2 K_w^2 \right]. 
\end{multline}
Taking the first derivative with respect to $K$ gives:
\begin{multline}
\frac{d \ \underset{t > t_0}{\min} \lbrace \E \left[ \| \nabla F(\bar{\bm{x}}_{t_0}) \|^2 \right] \rbrace}{d K_w} \\
= \frac{-2 \kappa (\kappa F(\bar{\bm{x}}_{t_0}) - F^*)}{\eta W K_w^2} \left[ \frac{|\bm{x}|}{D} + \frac{|\bm{x}|}{U} + \beta K_w \right] \\ 
+ 2 \eta \kappa L \left( 8 + \frac{4}{N} \right) G^2 K_w . 
\end{multline}
Taking the second derivative with respect to $K_w$ gives:
\begin{multline}
\frac{d^2 \ \underset{t > t_0}{\min} \lbrace \E \left[ \| \nabla F(\bar{\bm{x}}_{t_0}) \|^2 \right] \rbrace}{d K_w^2} \\
= \frac{4 \kappa (\kappa F(\bar{\bm{x}}_{t_0}) - F^*)}{\eta W K_w^3} \left[ \frac{|\bm{x}|}{D} + \frac{|\bm{x}|}{U} + \beta K_w \right] \\ 
+ 2 \eta \kappa L \left( 8 + \frac{4}{N} \right) G^2.
\end{multline}
Considering $(F(\bar{\bm{x}}_{t_0}) - F^*) > 0$ and all the constants in (25) are $> 0$, then inspection of (25) shows that the second derivative with respect to $K_w$ is greater than 0, and hence (23) is convex. Solving $\nicefrac{d \ \underset{t > t_0}{\min} \lbrace \E \left[ \| \nabla F(\bar{\bm{x}}_{t_0}) \|^2 \right] \rbrace}{d K_w} = 0$ gives Theorem 2.

\subsection{Proof of Corollary 2.1}
As with the proof of Theorem 2, we start with the bound on gradient norms using a constant $K$ and $\eta_w$ (within a communication round) given in (8):
\begin{multline}
\underset{t > t_0}{\min} \lbrace \E \left[ \| \nabla F(\bar{\bm{x}}_{t_0}) \|^2 \right] \rbrace \\
\leq \frac{2 \kappa (\kappa F(\bar{\bm{x}}_{t_0}) - F^*)}{\eta_w W K} \left[ \frac{|\bm{x}|}{D} + \frac{|\bm{x}|}{U} + \beta K \right] \\ 
+ \eta_w \kappa L \left[ \sum^{C}_{c = 1} p^2_c \sigma^2_c + 6 L \Gamma + \left( 8 + \frac{4}{N} \right) G^2 K^2 \right]. 
\end{multline}
Taking the first derivative with respect to $\eta_w$ gives: 
\begin{multline}
\frac{d \ \underset{t > t_0}{\min} \lbrace \E \left[ \| \nabla F(\bar{\bm{x}}_{t_0}) \|^2 \right] \rbrace}{d \ \eta_w} \\
= \frac{-2 \kappa (\kappa F(\bar{\bm{x}}_{t_0}) - F^*)}{\eta_w^2 W K} \left[ \frac{|\bm{x}|}{D} + \frac{|\bm{x}|}{U} + \beta K \right] \\ 
+ \kappa L \left[ \sum^{C}_{c = 1} p^2_c \sigma^2_c + 6 L \Gamma + \left( 8 + \frac{4}{N} \right) G^2 K^2 \right]. 
\end{multline}
Taking the second derivative with respect to $\eta_w$ gives: 
\begin{multline}
\frac{d^2 \ \underset{t > t_0}{\min} \lbrace \E \left[ \| \nabla F(\bar{\bm{x}}_{t_0}) \|^2 \right] \rbrace}{d \ \eta_w^2} \\
= \frac{4 \kappa (\kappa F(\bar{\bm{x}}_{t_0}) - F^*)}{\eta_w^3 W K} \left[ \frac{|\bm{x}|}{D} + \frac{|\bm{x}|}{U} + \beta K \right]. \\
\end{multline}
Noting that $(F(\bar{\bm{x}}_{t_0}) - F^*) > 0$ and all the constants in (28) are $> 0$, then inspection of (28) shows that the second derivative with respect to $\eta_w$ is $> 0$ and hence (27) is convex. Solving $\nicefrac{d \ \underset{t > t_0}{\min} \lbrace \E \left[ \| \nabla F(\bar{\bm{x}}_{t_0}) \|^2 \right] \rbrace}{d \eta_w} = 0$ yields Corollary 2.1.

\end{document}